%% file: acl_latex.tex
\documentclass[11pt]{article}

\usepackage[preprint]{acl}

\usepackage{times}
\usepackage{latexsym}

\usepackage[T1]{fontenc}

\usepackage[utf8]{inputenc}

\usepackage{microtype}

\usepackage{inconsolata}

\usepackage{graphicx}

\usepackage{amsmath}
\usepackage{booktabs}
\usepackage{newfloat}
\usepackage{listings}

\title{Conceptual Cultural Index: A Metric for Cultural Specificity \\ via Relative Generality}

\author{Takumi Ohashi \\
  Hosei University, Tokyo, Japan \\
  \texttt{takumi.ohashi.4g@gmail.com} \\\And
  Hitoshi Iyatomi \\
  Hosei University, Tokyo, Japan \\
  \texttt{iyatomi@hosei.ac.jp} \\}

\begin{document}
\maketitle

\begin{abstract}
    \input{00_abstract}
\end{abstract}

\section{Introduction}
    \input{01_introduction}
\section{Related Work}
    \input{02_relatedwork}
\section{CCI}
    \input{03_cci}
\section{Experiments}
    \input{04_experiments}
\section{Results and Discussion}
    \input{05_result_discussion}
\section{Conclusion}
    \input{06_conclusion}

\section*{Limitations}
    \input{limitations}


\bibliography{custom}

\appendix
    \input{appendix}

\end{document}

%% file: 00_abstract.tex
Large language models (LLMs) are increasingly deployed in multicultural settings; however, systematic evaluation of cultural specificity at the sentence level remains underexplored. 
We propose the Conceptual Cultural Index (CCI), which estimates cultural specificity at the sentence level. 
CCI is defined as the difference between the generality estimate within the target culture and the average generality estimate across other cultures. 
This formulation enables users to operationally control the scope of culture via comparison settings and provides interpretability, since the score derives from the underlying generality estimates.
We validate CCI on 400 sentences (200 culture-specific and 200 general), and the resulting score distribution exhibits the anticipated pattern: higher for culture-specific sentences and lower for general ones.
For binary separability, CCI outperforms direct LLM scoring, yielding more than a 10-point improvement in AUC for models specialized to the target culture.
Our code is available at \url{https://github.com/IyatomiLab/CCI}.

%% file: 01_introduction.tex
Large language models (LLMs) exhibit broad multilingual competence and are increasingly used for tasks such as search, summarization, and dialogue~\cite{gpt2020, gpt42023, llama2024, qwen2024}.
However, as applications expand, a key challenge remains in determining whether it is possible to ensure consistent and culturally aware responses across different regions. 
Everyday knowledge, such as dietary practices, greeting conventions, linguistic expressions, and seasonal events, varies systematically across cultures and regions, and how models handle this knowledge has direct implications for fairness, safety, and reliability~\cite{assessing2023, beer2024, understanding2024, normad2025}. 
Consequently, there is a need to develop models that can appropriately accommodate diverse cultural characteristics and differences, as well as models specialized for a specific culture.

Benchmarks of LLM capabilities have progressed beyond general-knowledge evaluations toward frameworks that focus on cultural knowledge, which advance the visualization of regional differences and biases~\cite{blend2024, culturalbench2025}. 
Many benchmarks use QA formats with overall accuracy as the main metric, consistently showing that culturally specific questions are harder than culture-agnostic ones~\cite{understanding2024, calmqa2025}.
However, cultural knowledge spans a continuum—from phenomena shared across regions to those unique to a specific locale—and existing benchmarks fail to capture this aspect, making it difficult to conduct error analysis and formulate targeted improvement strategies.

On the data side, large-scale corpora of cultural knowledge have been proposed~\cite{candle2023, culturebank2024}, but they lack annotations indicating the degree of cultural specificity of each sentence.
Thus, both evaluation and data resource development require a framework for quantitatively assessing sentence-level cultural specificity, yet no such framework currently exists.
Manual annotation of sentence-level cultural specificity is labor-intensive, requires domain expertise and contextual understanding, and often yields low inter-annotator agreement, highlighting the need for automation.
However, culture is a multifaceted, high-level construct, and prior work has rarely provided an explicit definition~\cite{adilazuarda2024}. 

In this paper, we propose the Conceptual Cultural Index (CCI), a sentence-level metric for quantifying cultural specificity, to address this challenge. 
CCI uses an LLM to estimate a sentence’s generality across multiple cultures and, based on these scores, quantifies the target culture’s specificity relative to others. 
This formulation allows users to control the scope of ``culture'' by adjusting the set of non-target cultures used for comparison.

The contributions of this study are as follows:
\begin{itemize}
  \item We introduce CCI, a new sentence-level metric for quantifying cultural specificity.
  \item Compared with direct LLM-based scoring of cultural specificity, CCI yields clearer separability between culture-specific and general sentences and offers greater interpretability.
  \item We present a practical use case of CCI by assigning item-level CCI scores to existing benchmarks and showing that model performance varies with the level of cultural specificity, enabling culture-aware error analysis.
\end{itemize}

%% file: 02_relatedwork.tex
Cultural evaluation benchmarks for language models include broad, multi-region datasets such as GeoMLAMA~\cite{geomlama2022}, BLEnD~\cite{blend2024}, CDEval~\cite{cdeval2024}, and CulturalBench~\cite{culturalbench2025}, as well as country- or region-specific benchmarks such as CLIcK~\cite{click2024}, IndoCulture~\cite{indoculture2024}, and CHARM~\cite{charm2024}. 
These resources support comparisons of cultural knowledge across models, but most adopt a QA format and rely on overall accuracy, which conflates culture-specific difficulty with general knowledge errors.

Text-based resources for collecting cultural knowledge, including StereoKG~\cite{stereokg2022}, CANDLE~\cite{candle2023}, CultureAtlas~\cite{cultureatlas2024}, CultureBank~\cite{culturebank2024}, and MANGO~\cite{mango2024}, extract and organize cultural assertions at scale.
However, they do not provide sentence-level annotations with continuous scores of cultural specificity, leaving a gap that our sentence-level metric CCI aims to fill.

%% file: 03_cci.tex
We propose the Conceptual Cultural Index (CCI), a sentence-level index of cultural specificity.
As illustrated in Figure~\ref{fig:cci}, given a target culture and a set of comparison cultures, we use an LLM to estimate how common a sentence is in each culture and derive a specificity score for the target culture.

\input{figures/tex/method_cci}

\subsection{Obtaining Generality Scores}
Given an input sentence \(x\), a set of cultures \(C\), and a target culture \(t \in C\), we use an LLM\footnote{Model selection is discussed in the experiments section.} to estimate, for each \(c \in C\), how common \(x\) is in the culture \(c\), yielding a continuous generality score \(p_c(x) \in [0,1]\).
In practice, we query all cultures in \(C\) within a single prompt and parse the scores from a JSON-formatted response.
The prompt used for the generality score is provided in Appendix~\ref{apd:prompt}.

To mitigate run-to-run variability in LLM outputs, we average results over \(N\) independent runs (in this paper, \(N=3\)):
\begin{equation}
  \bar p_c(x)
  = \frac{1}{N}\sum_{n=1}^{N} f_{\mathrm{LLM}}^{(n)}(x;C)[c],
  \quad c \in C.
\end{equation}
Here, \(f_{\mathrm{LLM}}^{(n)}(x;C)[c]\) denotes the score for culture \(c\) returned by the \(n\)-th run.

\subsection{Definition of CCI}
For a target culture \(t \in {C}\), we define CCI as the difference between the generality score in the target culture and the average generality score across the other cultures:
\begin{equation}
\label{eq:cci}
  CCI(x;t,{C})
  = \bar p_t(x)
    - \frac{1}{|{C}|-1}
      \sum_{c\in{C}\setminus\{t\}} \bar p_c(x).
\end{equation}
CCI takes values in \([-1,1]\): values near \(0\) indicate that \(x\) is cross-culturally general, values near \(1\) indicate that \(x\) is specific to the target culture, and values near \(-1\) indicate that \(x\) is specific to non-target cultures.

For Eq.~(\ref{eq:cci}), we also examined a sharpness-based formulation that weights the target culture by its log-softmax-normalized generality\footnote{$q_t=\dfrac{\exp(\bar p_t)}{\sum_{c\in C}\exp(\bar p_c)}$, 
$CCI_{\log}=\Bigl(1+\dfrac{\log q_t}{\log(|C|)}\Bigr)\,\bar p_t$.}, but it produced nearly constant scores across input sentences and compressed the score range, especially for larger \(|C|\).
In contrast, the simple difference is less sensitive to \(|C|\) and directly measures the gap on the original \([0,1]\) scale, so we adopt this definition.

%% file: figures/tex/method_cci.tex
\begin{figure}[t]
 \centering
  \includegraphics[width=1.0\linewidth]{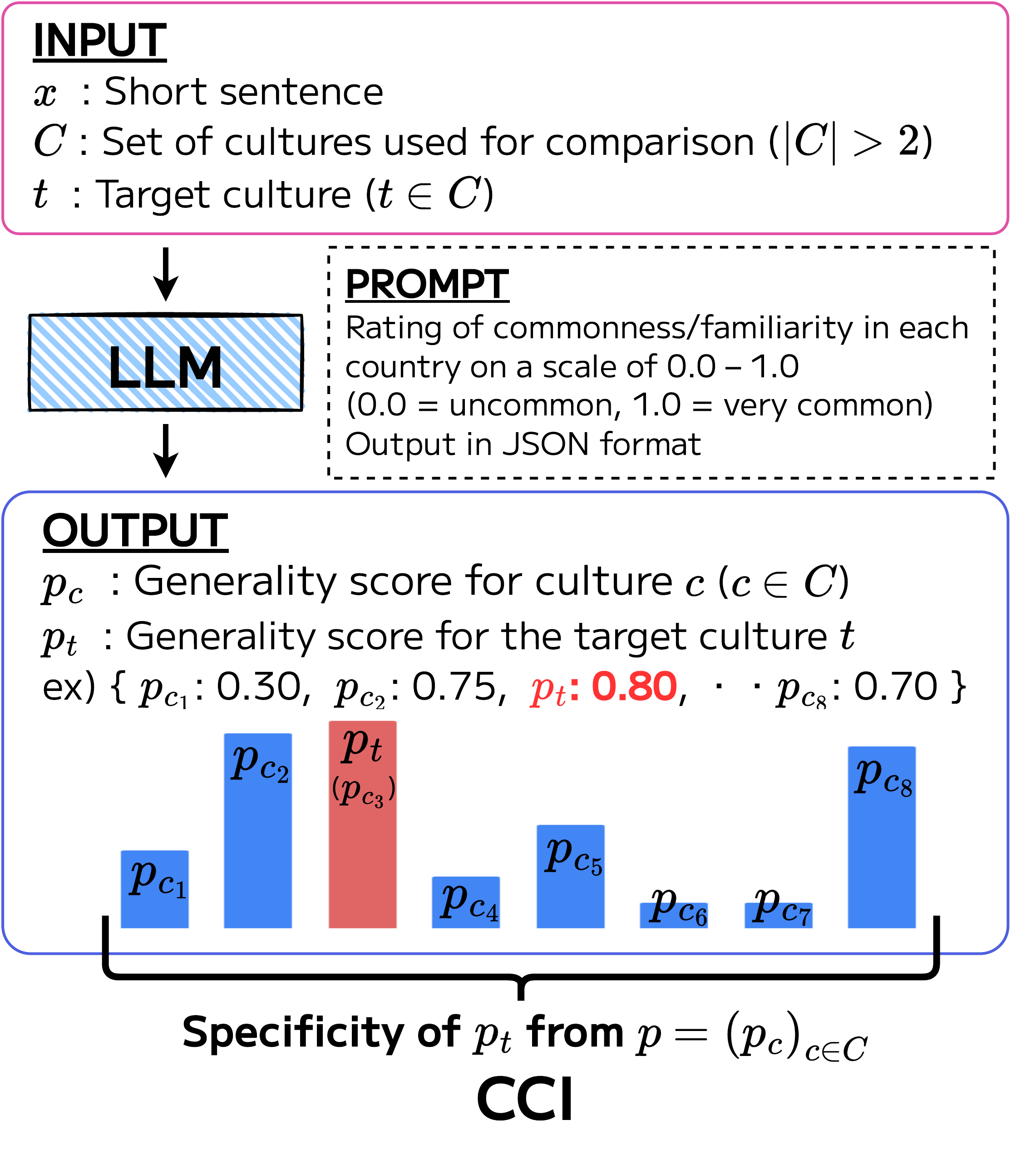}
  \caption{Overview of CCI.}
  \label{fig:cci}
\end{figure}

%% file: 04_experiments.tex
\subsection{Experimental Setup}
To assess whether CCI reflects cultural specificity, we use Japan as the target culture \(t\) and compute CCI for two classes: culture-specific sentences (positive) and general sentences (negative). 
We plot ROC curves for detecting Japanese cultural sentences and evaluate separability using the area under the ROC curve (AUC) and the difference in class medians.
We also conduct a qualitative analysis of representative examples.

As a baseline, we use an LLM that directly outputs a \([0,1]\) specificity score and compute AUC with the same protocol to compare its discriminative performance with CCI.
As with CCI, 0 denotes cross-cultural generality and 1 denotes specificity to the target culture.
The prompt used for the baseline is provided in Appendix~\ref{apd:prompt}.

\noindent\textbf{Data~~}
We first used GPT-5 (model as of August 7, 2025) to generate 300 short Japanese cultural sentences and 300 general sentences.
All generated sentences were manually reviewed and filtered, with duplicates and clear misclassifications removed, although some borderline cases may remain because cultural boundaries are inherently ambiguous.
The final evaluation set consists of 200 Japanese cultural sentences and 200 general sentences.
The prompts used for data generation are provided in Appendix~\ref{apd:prompt}.

\noindent\textbf{Models~~}
To identify suitable LLMs for computing CCI, we compared CCI and the baseline under a common protocol across five LLMs: multilingual models (Llama~3.1~\cite{llama2024}, Qwen~2.5~\cite{qwen2024}, gpt-oss~\cite{gptoss2025})
and Japanese-specialized models (Llama~3.1 Swallow~\cite{swallow2024}, llm-jp~3.1~\cite{llmjp2024}).
The exact model identifiers and links are listed in Appendix~\ref{apd:model}.

\subsection{Varying the Set of Cultures \(C\)}
CCI allows the set \(C\) of cultures to vary.
To assess how controllable the cultural scope is, we conduct experiments under two modes.

\noindent\textbf{Global mode~~}
\(C\) is fixed to the 19 G20 member countries, excluding the European Union and the African Union, as we restrict \(C\) to country names.

\noindent\textbf{Custom mode~~}
\(C\) is configured to match the task objective. In this experiment, to test whether the inclusion of neighboring cultures can be controlled, we defined two conditions across four countries:
\begin{enumerate}
\item \textbf{\emph{+Neighbor Culture}}, which includes neighboring countries in \(C\): 
[``China'', ``Republic of Korea'', ``United States of America'', ``Japan''];\,
\item \textbf{\emph{--Neighbor Culture}}, which excludes neighboring countries from \(C\): 
[``Brazil'', ``France'', ``United States of America'', ``Japan''];\,
\end{enumerate}

For the baseline, we also evaluate two prompting conditions: one prompt that explicitly includes the instruction
\emph{``If the practice is also common in neighboring or culturally adjacent countries, do not consider it specific to the target.''}
(\textbf{\emph{+Neighbor Culture}}), and another that omits this instruction.

\subsection{CCI-Based Benchmark Stratification}
As a use case of CCI, we perform benchmark stratification by assigning CCI scores to each item and analyzing how task accuracy varies across CCI levels.
We use two datasets that capture Japanese commonsense: JCommonsenseQA (JCQA)~\cite{jglue2022}\footnote{\url{https://github.com/yahoojapan/JGLUE}} and JCommonsenseMorality (JCM)~\cite{jethics2025}\footnote{\url{https://github.com/Language-Media-Lab/jethics}}, both of which include items that may reflect phenomena specific to the Japanese cultural sphere.
JCQA is a five-way multiple-choice commonsense question answering task; we compute CCI using as input \(x\) the concatenation of the question text and the gold option text.
JCM is a binary classification task that judges whether an action is morally acceptable; we compute CCI using as input \(x\) the target sentence together with the gold label.
For scoring, we use CCI in Global mode computed with gpt-oss.
We evaluate on the JCQA dev set (1{,}119 items) and the JCM test set (3{,}992 items) using predictions from Qwen~2.5, Llama~3.1, and llm-jp~3.1.

\input{tables/tex/result_median}
\input{tables/tex/result_auc}

%% file: tables/tex/result_median.tex
\begin{table*}[t]
  \centering
  \setlength{\tabcolsep}{1.5pt}
  \begin{small}
    \resizebox{2.05\columnwidth}{!}{
    \begin{tabular}{@{}p{3.0cm}ccccc@{}} 
      \toprule
      Models &
        \multicolumn{2}{c}{Baseline} &
        CCI (Global) &
        \multicolumn{2}{c}{CCI (Custom)} \\
      \cmidrule(lr){2-3}\cmidrule(lr){4-4}\cmidrule(lr){5-6}
      &
        \shortstack{($C_{\mathrm{median}}\!\uparrow$, $G_{\mathrm{median}}\!\downarrow$)} &
        \shortstack{+Neighbor\\ ($C_{\mathrm{median}}\!\uparrow$, $G_{\mathrm{median}}\!\downarrow$)} &
        \shortstack{($C_{\mathrm{median}}\!\uparrow$, $G_{\mathrm{median}}\!\downarrow$)} &
        \shortstack{+Neighbor\\ ($C_{\mathrm{median}}\!\uparrow$, $G_{\mathrm{median}}\!\downarrow$)} &
        \shortstack{-Neighbor\\ ($C_{\mathrm{median}}\!\uparrow$, $G_{\mathrm{median}}\!\downarrow$)} \\
      \midrule
      Qwen2.5-7B               & (0.815, 0.800) & (0.980, 0.800) & (0.800, 0.505) & (0.633, 0.267) & (0.833, 0.467) \\
      Llama-3.1-8B             & (0.870, 0.800) & (0.870, 0.800) & (0.778, 0.648) & (0.664, 0.283) & (0.980, 0.711) \\
      Llama-3.1-Swallow-8B & (0.950, 0.850) & (0.950, 0.850) & (0.761, 0.324) & (0.331, 0.117) & (0.933, 0.300) \\
      llm-jp-3.1-13b          & (0.800, 0.785) & (0.800, 0.700) & (0.869, 0.568) & (0.792, 0.467) & (0.897, 0.593) \\
      gpt-oss-20b                        & (0.880, 0.100) & (0.775, 0.100) & (0.836, 0.063) & (0.697, 0.111) & (0.817, 0.104) \\
      \bottomrule
    \end{tabular}
  }
  \end{small}
  \caption{Class-wise medians of specificity scores for CCI and the baseline. \(C_{\mathrm{median}}\) denotes the median score for culture-specific sentences, and \(G_{\mathrm{median}}\) denotes the median score for general sentences. While \(C_{\mathrm{median}}\) should ideally be close to 1 and \(G_{\mathrm{median}}\) close to 0, it is not necessary for every instance to reach these extremes; it suffices that the overall trend is observed. The cultural specificity score aims to assign context-appropriate values to each sentence.}
  \label{tbl:result_median}
\end{table*}

%% file: tables/tex/result_auc.tex
\begin{table}[t]
  \centering
  \setlength{\tabcolsep}{5pt}
  \begin{small}
  \resizebox{\columnwidth}{!}{%
  \begin{tabular}{lcc}
    \toprule
    Models &
      \shortstack{Baseline\\ AUC / \(\Delta\)} &
      \shortstack{CCI (Global)\\ AUC / \(\Delta\)} \\
    \midrule
    Qwen2.5-7B         & 0.816 / 0.015 & \textbf{0.884} / \textbf{0.295} \\
    Llama-3.1-8B       & \textbf{0.803} / 0.070 & 0.796 / \textbf{0.130} \\
    Llama-3.1-Swallow-8B  & 0.842 / 0.100 & \textbf{0.945} / \textbf{0.437} \\
    llm-jp-3.1-13b    & 0.768 / 0.015 & \textbf{0.908} / \textbf{0.301} \\
    gpt-oss-20b                 & \textbf{0.963} / \textbf{0.780} & 0.956 / 0.773 \\
    \bottomrule
  \end{tabular}
  }
  \end{small}
  \caption{Separability between culture-specific and general sentences, reported in terms of the AUC and the median gap \(\Delta = C_{\mathrm{median}} - G_{\mathrm{median}}\).}
  \label{tbl:result_auc}
\end{table}

%% file: 05_result_discussion.tex
\subsection{Separability between Culture-Specific and General Sentences}
Table~\ref{tbl:result_median} shows, for each LLM, the class-wise median specificity scores for CCI and the direct-estimation baseline: \(C_{median}\) for culture-specific sentences and \(G_{median}\) for general sentences.
Table~\ref{tbl:result_auc} shows separability, measured by AUC, and the gap between the medians \((\Delta = C_{median} - G_{median})\).
We report median gaps rather than mean gaps because the baseline scores are bounded in \([0,1]\), whereas CCI scores lie in \([-1,1]\).
Given these different ranges, mean differences could inadvertently favor CCI; using medians avoids this issue in our experiments.

\noindent\textbf{CCI vs. Baseline~~}
CCI achieves AUC comparable to or higher than the baseline and yields clearer separation, with higher scores for culture-specific sentences and lower scores for general ones.
The baseline, in contrast, tends to assign relatively high scores to many sentences, and for some models the class medians are nearly identical.
This may be because directly quantifying ``culture'' as a single scalar is inherently difficult, whereas CCI decomposes the task into per-culture generality estimates, thereby stabilizing the inference process.

\noindent\textbf{Model suitability~~}
Regarding which LLMs are suitable for computing cultural specificity scores, gpt-oss achieves near-ideal separation under both CCI and the baseline.
This appears to reflect not only model size but also its reasoning-oriented architecture, which captures cultural differences through step-by-step reasoning.
In addition, Japanese-specialized models show better separation than multilingual models.
Overall, models that combine strong reasoning capabilities and a deep understanding of the target culture, while also possessing knowledge of other cultures, are most suitable for computing cultural specificity scores.

\input{figures/tex/result_example}

\subsection{Controllability of Cultural Scope}
\noindent\textbf{CCI vs. Baseline~~}
From Table~\ref{tbl:result_median}, we observe that under CCI’s Custom mode (\emph{+Neighbor}), the median score for the culture-specific class is lower than in the Global mode.
This suggests that the cultural scope can be adjusted to avoid overestimating practices common in neighboring cultures.
By contrast, the baseline appears to be sensitive to prompt wording and input language, indicating that cultural scope is difficult to control solely through textual instructions.
Additionally, CCI provides a numerical assessment of cultural specificity together with per-culture generality scores; even when baseline accuracy is high, CCI offers greater interpretability by indicating in which cultures a sentence is considered common or uncommon.

\noindent\textbf{Case analysis~~}
Figure~\ref{fig:result_ex} shows a subset of the evaluation instances along with the CCI scores produced by gpt-oss.
In individual cases, actions that may be taboo in neighboring cultures (e.g., ``Pick up the small bowl and bring it to your mouth.'') should not receive excessively low scores even under \emph{+Neighbor}.
Conversely, practices widely observed across regions (e.g., ``Taking milk out of the refrigerator.'') should not receive high scores even under \emph{--Neighbor}.
Consistent with these expectations, the presence or absence of similar practices in neighboring cultures yields systematic differences between the \emph{+Neighbor} and \emph{--Neighbor} conditions, indicating that CCI effectively operationalizes cultural scope control as intended.

\input{tables/tex/result_jcqa}
\input{tables/tex/result_jcm}

\subsection{Task Accuracy Shifts across CCI Levels}
Tables~\ref{tbl:result_jcqa} and~\ref{tbl:result_jcm} show the results for JCQA and JCM, respectively, where items are binned by CCI in increments of 0.1.
Across both datasets, the CCI distribution is skewed toward lower values, suggesting that JCQA and JCM contain many culturally non-specific commonsense questions.

Accuracy tends to decrease as CCI increases, and higher-CCI bins often fall below the overall dataset accuracy.
This indicates that items with higher cultural specificity tend to be more challenging, and that current models may not sufficiently acquire culture-specific knowledge.
In contrast, llm-jp maintains higher overall accuracy and exhibits a comparatively smaller drop in high-CCI bins.
This suggests that models trained on Japanese data may have an advantage on items that can reflect commonsense knowledge in the Japanese cultural sphere.
Overall, CCI-based stratification makes clear how performance varies with cultural specificity. This variation is difficult to observe from overall accuracy alone, and the results suggest that model gaps tend to widen in higher-CCI bins.

%% file: figures/tex/result_example.tex
\begin{figure*}[t]
 \centering
  \includegraphics[width=1.0\linewidth]{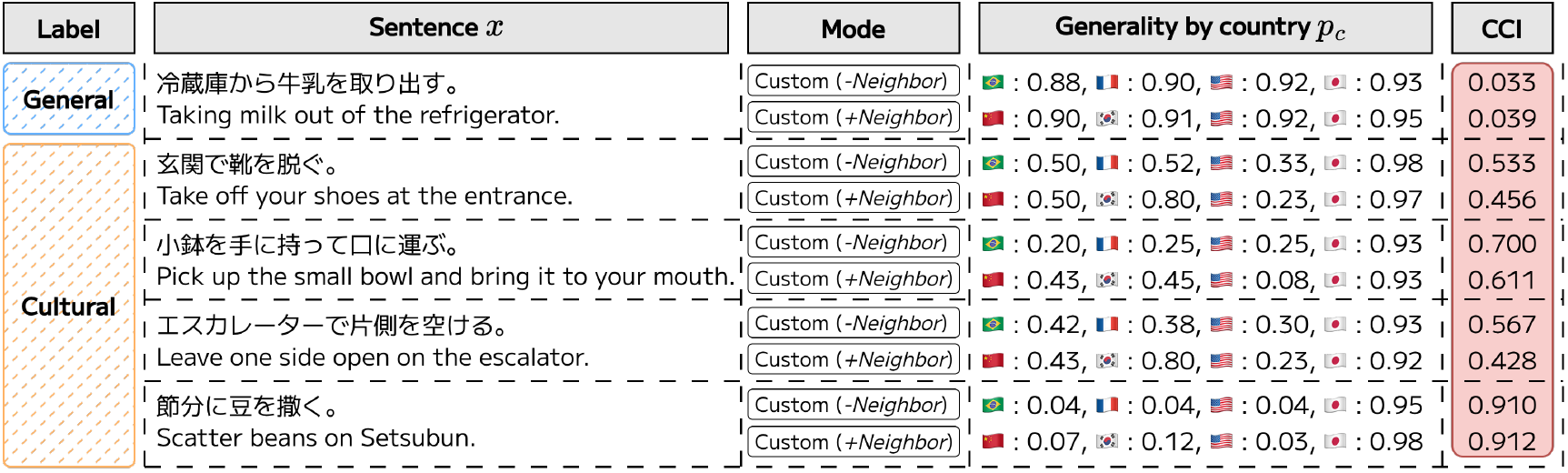}
  \caption{Example sentences and their corresponding CCI scores computed by gpt-oss.}
  \label{fig:result_ex}
\end{figure*}

%% file: tables/tex/result_jcqa.tex
\begin{table}[t]
  \centering
  \setlength{\tabcolsep}{2pt}
  \begin{small}
  \resizebox{\columnwidth}{!}{%
  \begin{tabular}{l r @{\hspace{10pt}} c c c}
    \toprule
    \multicolumn{2}{c}{Bin statistics} & \multicolumn{3}{c}{Accuracy} \\
    \cmidrule(r){1-2} \cmidrule(l){3-5}
    Range & \#Items & Qwen~2.5 & Llama~3.1 & llm-jp~3.1 \\
    \midrule
    CCI $\le$ 0.1              & 583 & 0.940 & 0.899 & 0.967 \\
    0.1 $<$ CCI $\le$ 0.2      & 75  & 0.893 & 0.880 & 0.987 \\
    0.2 $<$ CCI $\le$ 0.3      & 69  & 0.826 & 0.870 & 0.971 \\
    0.3 $<$ CCI $\le$ 0.4      & 40  & 0.900 & 0.775 & 0.925 \\
    0.4 $<$ CCI $\le$ 0.5      & 19  & 0.789 & 0.789 & 0.842 \\
    0.5 $<$ CCI $\le$ 0.6      & 63  & 0.873 & 0.825 & 0.905 \\
    0.6 $<$ CCI $\le$ 0.7      & 36  & 0.889 & 0.806 & 0.972 \\
    0.7 $<$ CCI $\le$ 0.8      & 45  & 0.867 & 0.800 & 0.978 \\
    0.8 $<$ CCI $\le$ 0.9      & 173 & 0.873 & 0.821 & 0.942 \\
    0.9 $<$ CCI $\le$ 1.0      & 16  & 0.750 & 0.750 & 0.938 \\
    \midrule
    \multicolumn{2}{l}{Overall Accuracy} & 0.904 & 0.864 & 0.958 \\
    \bottomrule
  \end{tabular}
  }
  \end{small}
  \caption{Number of JCQA items and model-wise accuracy when CCI is binned in increments of 0.1.}
  \label{tbl:result_jcqa}
\end{table}

%% file: tables/tex/result_jcm.tex
\begin{table}[t]
  \centering
  \setlength{\tabcolsep}{2pt}
  \begin{small}
  \resizebox{\columnwidth}{!}{%
  \begin{tabular}{l r @{\hspace{10pt}} c c c}
    \toprule
    \multicolumn{2}{c}{Bin statistics} & \multicolumn{3}{c}{Accuracy} \\
    \cmidrule(r){1-2} \cmidrule(l){3-5}
    Range & \#Items & Qwen~2.5 & Llama~3.1 & llm-jp~3.1 \\
    \midrule
    CCI $\le$ 0.1              & 3217 & 0.846 & 0.836 & 0.924 \\
    0.1 $<$ CCI $\le$ 0.2      & 117  & 0.778 & 0.795 & 0.906 \\
    0.2 $<$ CCI $\le$ 0.3      & 116  & 0.741 & 0.724 & 0.897 \\
    0.3 $<$ CCI $\le$ 0.4      & 72   & 0.708 & 0.694 & 0.931 \\
    0.4 $<$ CCI $\le$ 0.5      & 60   & 0.733 & 0.767 & 0.883 \\
    0.5 $<$ CCI $\le$ 0.6      & 75   & 0.680 & 0.773 & 0.880 \\
    0.6 $<$ CCI $\le$ 0.7      & 94   & 0.755 & 0.691 & 0.894 \\
    0.7 $<$ CCI $\le$ 0.8      & 138  & 0.761 & 0.681 & 0.891 \\
    0.8 $<$ CCI $\le$ 0.9      & 102  & 0.745 & 0.657 & 0.892 \\
    0.9 $<$ CCI $\le$ 1.0      & 1    & 1.000 & 1.000 & 1.000 \\
    \midrule
    \multicolumn{2}{l}{Overall Accuracy} & 0.826 & 0.814 & 0.918 \\
    \bottomrule
  \end{tabular}
  }
  \end{small}
  \caption{Number of JCM items and model-wise accuracy when CCI is binned in increments of 0.1.}
  \label{tbl:result_jcm}
\end{table}

%% file: 06_conclusion.tex
We introduced the Conceptual Cultural Index (CCI), a metric that quantifies sentence-level cultural specificity as the relative difference in generality between a target culture and other cultures.
CCI remains effective even when the baseline struggles to assign stable scores and provides interpretable estimates grounded in an explicit definition.
CCI supports practical culture-related workflows, including annotating and stratifying benchmarks for model evaluation, as well as filtering culture-specific knowledge data.
Our method is presented as an evaluation framework that can be applied consistently as models evolve.

%% file: limitations.tex
This study has three limitations.
First, because cultures were approximated primarily at the country level, intra-country heterogeneity, such as regional or generational differences, may not be fully captured. 
Second, our experiments focused on Japan as the target culture, and thus the generalizability to other languages and regions remains to be established.
Third, CCI relies on LLM-generated generality scores and thus inherits the biases and calibration issues of the underlying models. 
With these points in mind, we plan to develop CCI into a more robust and general-purpose evaluation framework by refining the granularity of cultural groupings and conducting broader multilingual and cross-regional evaluations.

%% file: appendix.tex
\section{Prompts}
\label{apd:prompt}
This section provides the prompts used in our experiments.
Table~\ref{tbl:appendix_generality} presents the prompt for obtaining per-culture generality scores used to compute CCI; Table~\ref{tbl:appendix_compare} shows the prompt for direct cultural specificity scoring used in the baseline comparison; and Table~\ref{tbl:appendix_generate_general} and ~\ref{tbl:appendix_generate_cultural} provides the prompts used to generate the evaluation data (culture-specific and general sentences).

\input{tables/tex/appendix_generality}
\input{tables/tex/appendix_compare}
\input{tables/tex/appendix_generate_general}
\input{tables/tex/appendix_generate_cultural}

\section{Model Details}
\label{apd:model}
We use the following LLMs in our experiments:
\begin{itemize}
  \item Llama 3.1 8B Instruct:\\ \url{https://huggingface.co/meta-llama/Llama-3.1-8B-Instruct}
  \item Qwen 2.5 7B Instruct:\\ \url{https://huggingface.co/Qwen/Qwen2.5-7B-Instruct}
  \item gpt-oss-20B:\\ \url{https://huggingface.co/openai/gpt-oss-20b}
  \item Llama 3.1 Swallow 8B Instruct v0.5:\\ \url{https://huggingface.co/tokyotech-llm/Llama-3.1-Swallow-8B-Instruct-v0.5}
  \item LLM-jp 3.1 13B instruct4:\\ \url{https://huggingface.co/llm-jp/llm-jp-3.1-13b-instruct4}
\end{itemize}

%% file: tables/tex/appendix_generality.tex
\begin{table*}[t]
  \centering
  \begin{tabular}{@{}p{\linewidth}@{}}
  \toprule
  \begin{lstlisting}
Task:
Rate how COMMON/FAMILIAR the following item is in each country (0.00 = not common, 1.00 = very common). Treat countries independently. Be language-agnostic: interpret the statement regardless of its language.

Statement: {sentence}
Countries: {cultures}

Rules:
- Use general knowledge; avoid stereotypes.
- If similarly common across many countries, use similar (even identical) scores.
- If unsure, use mid values (e.g., 0.50).
- Do NOT normalize across countries.

Output JSON ONLY (no prose):
Schema: {"scores": {"<country>": <float>}}
Constraints: use the country names exactly as provided; floats in [0.00, 1.00], rounded to two decimals.
  \end{lstlisting}
  \\
  \bottomrule
  \end{tabular}
  \caption{Prompt for obtaining per-culture generality scores used in computing CCI. The input consists of a sentence $x$ and a set of cultures $\mathcal{C}$.}
  \label{tbl:appendix_generality}
\end{table*}

%% file: tables/tex/appendix_compare.tex
\begin{table*}[t]
  \centering
  \begin{tabular}{@{}p{\linewidth}@{}}
  \toprule
  \begin{lstlisting}
Task:
Decide whether the following item is culturally specific to the TARGET country, and output a numeric specificity score only
(0.00 = globally common; 1.00 = unique to the target).

Statement: {sentence}
Target country: {target_culture}

Rules:
- Use general knowledge; avoid stereotypes.
- Be language-agnostic; interpret the statement regardless of its language.
- If unsure, use mid values (e.g., 0.50).

Output JSON ONLY (no prose):
{"score": <float in [0.00,1.00] rounded to two decimals>}
  \end{lstlisting}
  \\
  \bottomrule
  \end{tabular}
  \caption{Prompt for directly predicting cultural specificity scores in the target culture by an LLM (Baseline). The input consists of a sentence $x$ and a target culture $t$.}
  \label{tbl:appendix_compare}
\end{table*}

%% file: tables/tex/appendix_generate_general.tex
\begin{table*}[t]
  \centering
  \begin{tabular}{@{}p{\linewidth}@{}}
  \toprule
  \begin{lstlisting}
You are an assistant for creating a short-sentence corpus.
Strictly satisfy the requirements below.

Goal:
- Collect very ordinary events that could occur in any region of the world.

Output:
- Exactly one natural Japanese sentence, short in length (about 10-20 characters).
- Describe facts plainly without evaluations, impressions, or subjectivity.
- Avoid place names, specific store names, and personal names. Avoid excessive stereotypes.
- Vary expressions, vocabulary, and scenes so that the same sentence endings and the same constructions do not appear consecutively.

Strict requirements:
- The output must be a JSON array. Each element must have the form { "text": "<one sentence>" }.
- The number of items must be exactly 300.
- Do not include any additional explanations, labels, or numbering (do not output any strings other than JSON).

Example (format only):
[
  { "text": "Turn off the alarm in the morning." },
  { "text": "Wait for the train at the station." }
]
  \end{lstlisting}
  \\
  \bottomrule
  \end{tabular}
  \caption{Prompt for generating general sentences. We used a Japanese prompt in our experiments; the version shown here is the English translation.}
  \label{tbl:appendix_generate_general}
\end{table*}

%% file: tables/tex/appendix_generate_cultural.tex
\begin{table*}[t]
  \centering
  \begin{tabular}{@{}p{\linewidth}@{}}
  \toprule
  \begin{lstlisting}
You are an assistant for creating a short-sentence corpus.
Strictly satisfy the requirements below.

Goal:
- Broadly collect, in short sentences, Japan-specific customs, daily life culture, annual events, food culture, public manners, etc.

Output:
- Exactly one natural Japanese sentence, short in length (about 10-20 characters).
- Describe events plainly without evaluations or impressions.
- Avoid place names, specific store names, and personal names. Avoid excessive stereotypes.
- Vary expressions, vocabulary, and scenes so that the same sentence endings and the same constructions do not appear consecutively.

Strict requirements:
- The output must be a JSON array. Each element must have the form { "text": "<one sentence>" }.
- The number of items must be exactly 300.
- Do not include any additional explanations, labels, or numbering (do not output any strings other than JSON).

Example (format only):
[
  { "text": "Take off your shoes at the entrance." },
  { "text": "Use the purification basin at a shrine." }
]
  \end{lstlisting}
  \\
  \bottomrule
  \end{tabular}
  \caption{Prompt for generating Japan-specific (culture-specific) sentences. We used a Japanese prompt in our experiments; the version shown here is the English translation.}
  \label{tbl:appendix_generate_cultural}
\end{table*}